\documentclass[letterpaper]{article}
\usepackage{amsmath,amssymb,amsfonts}
\usepackage{algorithm2e}
\usepackage{graphicx}
\usepackage{textcomp}
\usepackage[dvipsnames]{xcolor}
\usepackage{tikz}
\usepackage{color}

\DeclareMathOperator*{\argmax}{argmax}
\def \sign {{\rm sign}}
\begin{document}

\title{Hierarchical Classification using Binary Data}

 \author{Denali Molitor \quad and \quad Deanna Needell \\
 University of California, Los Angeles \\ Los Angeles, CA 90095, USA}

\date{}
\maketitle

\begin{abstract}
In classification problems, especially those that categorize data into a large number of classes, the classes often naturally follow a hierarchical structure. That is, some classes are likely to share similar structures and features. Those characteristics can be captured by considering a hierarchical relationship among the class labels. Here, we extend a recent simple classification approach on binary data in order to efficiently classify hierarchical data. In certain settings, specifically, when some classes are significantly easier to identify than others, we showcase computational and accuracy advantages.
\end{abstract}

\section{Introduction}
We consider the problem of classification, where one is given a set of labeled data used for training, and from that data wishes to accurately assign labels to new unlabeled data. In the general problem, the class labels themselves have no relation to one another, however, data can often be organized in a hierarchical way. For example, in image classification problems, the data may contains images of inanimate and living objects. Then, within each of those classes the data may be further identified as images of vehicles and toys, or humans and animals.  The data could then be further subdivided into classes of various animal types, and so on. This structure can be visualized as a \textit{tree}, where the children of each node correspond to its sub-classes. Each data point in this case would have a label corresponding to a leaf of the tree, but also possesses the characteristics of all the labels of its ancestors. One option of course would be to simply use generic classification schemes to classify the data using the leaf labels only. \textit{Hierarchical classification}, however, makes use of information and structure between groups in classifying the data \cite{gordon1987review,silla2011survey}. Extensions of popular classification methods such as the support vector machine (SVM) to the hierarchical setting are not straightforward, and such approaches often decompose the problem into many sub-problems leading to higher computational complexities \cite{cheong2004support,weston1998multi}.

Recently, \cite{NSW17Simple} proposed a simple classification scheme that uses only binary representations of data to perform classification; such representations arise naturally or are particularly efficient in many applications, see e.g. \cite{fang2014sparse,biht,aziz1996overview,RefWorks:526,RefWorks:439}. Here, we show that this method lends itself well to performing hierarchical classification and, in particular, using the hierarchical structure to improve computational efficiency. The classification method uses position of data relative to random hyperplanes to predict in which class a point is most likely to belong.  \cite{NSW17Simple} demonstrated that for more complex data, using combinations of hyperplanes enables one to make more accurate predictions. However, the computation required to make a prediction scales exponentially in the number of hyperplane combinations used. Fortunately, the method is highly adjustable and for data that is likely to be more or less difficult to classify, one can adjust the number of these hyperplane combinations. Such a method is likely to be particularly useful for hierarchical data in which certain subclasses of data are more or less difficult to classify than others.

\section{Underlying Classification Algorithm}
In this section, we describe the classification algorithm proposed in  \cite{NSW17Simple}. Let $A$ be a random matrix in $\mathbb{R}^{m\times n}$, $X =[x_1\, x_2\, \cdots\, x_p]\in \mathbb{R}^{n\times p},$ where the $x_i\in\mathbb{R}^p$ are the data with labels $b=(b_1\,\cdots\, b_p)$ assigned from a possibility of $G$ classes. Suppose we only have access to binary measurements of the data in the form $Q = \sign(AX)$, where $\sign(M)_{i,j}=\sign(M_{i,j}$). The rows of $A$ will correspond to (random) hyperplanes, and thus $Q_{i,j}$ simply captures on which side of the $i$th hyperplane the $j$th data point lies.

\begin{figure}[ht]
\begin{center}
 \includegraphics[width=2in]{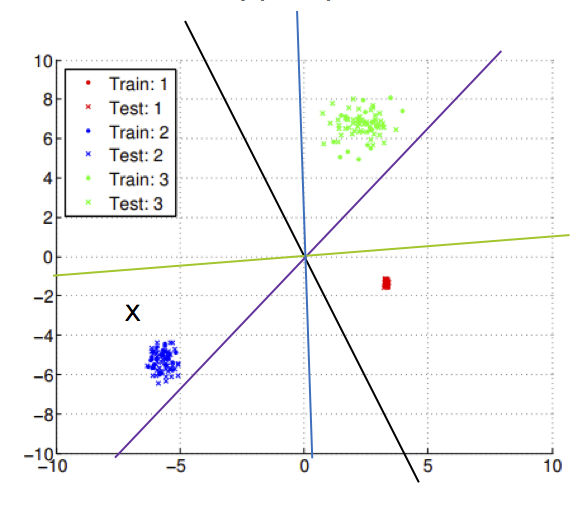} \quad \includegraphics[width=2in]{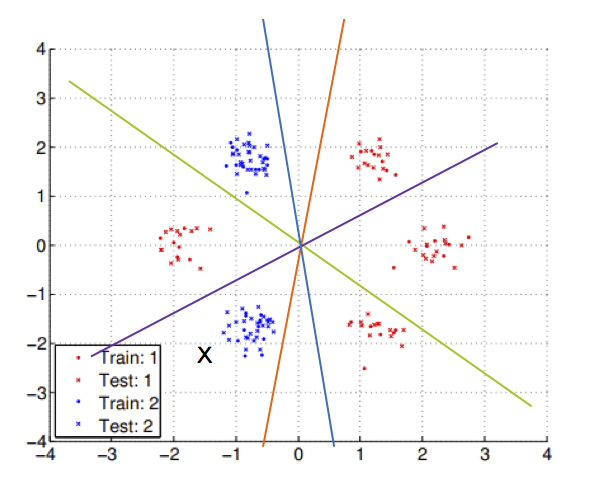}
\end{center}
\caption{Two motivating examples for the classification method.}\label{motiv}
\end{figure}

Let us build some intuition for the approach. Consider the two-dimensional data $X$ shown in the top plot of Figure \ref{motiv}, consisting of three labeled classes (green, blue, red). Consider the four hyperplanes shown in the same plot, and suppose we had access only to the binary data $Q=\sign(AX)$, where $A$ contains the normals to each hyperplane as its rows. For the new test point $x$ (which by visual inspection should be labeled blue) and its binary data $q = \sign(Ax)$, one could simply cycle through the hyperplanes and decide which class $x$ matches most often. For example, for the hyperplane colored purple in the plot, $x$ has the same sign (i.e. lies on the same side) as the blue and green classes. For the black hyperplane, $x$ only matches the blue class, and so on. Then for this example, $x$ will clearly match the blue class most often, and we could assign it that label correctly. However, next consider the more complicated geometry given in the bottom plot, where the data consists of only two classes (red and blue), but they are now no longer linearly separable. This same strategy will no longer be accurate for the test point $x$. However, now instead of single hyperplanes, consider hyperplane \textit{pairs}, and ask which class label $x$ most often matches (note that in this context, by ``matches" we now mean that points lie in the same cone into which the hyperplanes divide the space). For example, for the pair of hyperplanes colored orange and green, $x$ matches both red and blue points, whereas for the pair of hyperplanes colored orange and purple, $x$ only matches the blue class. One could now cycle through all pairs, and again ask which class $x$ matches most often. For complicated data, we could aggregate such information across various \textit{levels}, where at level $l$ we consider $l$-tuples of hyperplanes in this way.

Let us now describe the method more formally. Consider $m$ $l$-tuples of hyperplanes of various lengths $l = 1,\cdots, L$. 
Define the randomly selected index sets $\Lambda_{l,i}$, where $\Lambda_{l,i}\subset[m]$, $|\Lambda_{l,i}|=l$ and each $\Lambda_{l,i}$ is unique. If we then isolate the rows of $Q$ contained in $\Lambda_{l,i}$ to form the $l\times p$ matrix $Q^{\Lambda_{l,i}}$, the columns of this matrix give the sign patterns of the data with respect to the hyperplanes in $\Lambda_{l,i}$. Let $T_{l,i}$ be the number of unique sign patterns, or equivalently columns of $Q^{\Lambda_{l,i}}$.
Based on these sign patterns, we then calculate the membership index parameter $r(l,i,t,g)$ for each $l$-tuple $i = 1,\cdots , m$, level $l = 1,\cdots,L$, unique sign pattern $t = 1,\cdots, T_{l,i}$ and class $g=1, \cdots, G$. Let $P_{g|t}$ be the number of data points in class $g$ with sign pattern $t$ and define:
\begin{equation}\label{RF}
r(l,i,t,g) := \frac{P_{g|t}}{\sum_{j=1}^G P_{j|t}}\frac{\sum_{j=1}^G |P_{g|t}-P_{j|t}|}{\sum_{j=1}^G P_{j|t}}.
\end{equation}
The first fraction in \eqref{RF} measures the proportion of data with sign pattern $t$ that belong to class $g$, and the second fraction is a balancing term.
This training process is also described in Algorithm \ref{algo:Train}.

\begin{algorithm}
    \SetKwInOut{Input}{input}
    \SetKwInOut{Select}{select}
    \SetKwInOut{Determine}{determine}
    \SetKwInOut{Compute}{compute}

    \Input{binary training data $Q$, training labels $b$, number of classes $G$, number of levels $L$.}
    \For{$l$ from 1 to $L$, $i$ from 1 to $m$}
      {
       \Select{Randomly select $\Lambda_{l,i}\subset [m],|\Lambda_{l,i}|=l$.}
	\Determine{Determine the $T_{l,i}\in \mathbb{N}$ unique column patterns in $Q^{\Lambda_{l,i}}$.}
	\For{$t$ from $1$ to $T_{l,i},$ $g$ from $1$ to $G$}{
		\Compute{Compute $r(l,i,t,g)$ as in Equation \ref{RF}.}
	}
      }
    \caption{Training from  \cite{NSW17Simple}}
	\label{algo:Train}
\end{algorithm}

Given $x\in \mathbb{R}^n$ and $q=\sign(Ax)$, we predict the class of $x$ as given in Algorithm \ref{algo:Class}. 
Intuitively, the membership values $r(l,i,t,g)$ indicate how likely a point with sign pattern $t$ is to lie in class $g$, given information from the $i$th $l$-tuple of hyperplanes. These are aggregated over all measurements $m$ and levels $l$, giving a likelihood that the point belongs to each class $g$.  The label assigned is simply the class $g$ corresponding to the largest value of $\tilde r$.

\begin{algorithm}
    \caption{Classification from  \cite{NSW17Simple}}
	\label{algo:Class}
    \SetKwInOut{Input}{input}
    \SetKwInOut{Init}{initialize}
    \SetKwInOut{Id}{identify}
    \SetKwInOut{Update}{update}
    \SetKwInOut{Scale}{scale}
    \SetKwInOut{Classify}{classify}
    \SetKwInOut{Set}{set}

    \Input{binary testing data $q$, number of classes $G$, number of levels $L$, learned parameters $r(l,i,t,g)$, $T_{l,i}$, and $\Lambda_{l,i}$ from Algorithm~\ref{algo:Train}.}
    \For{$l$ from 1 to $L$, $i$ from 1 to $m$}
      {
       \Id{Identify the pattern $t^*\in [T_{l,i}]$ to which $q^{\Lambda_{l,i}}$ corresponds.}
	\For{$g$ from $1$ to $G$}{
		\Update{$\tilde r(g)=\tilde r(g)+r(l,i,t^*,g)$.}
	}
	}
	\Scale{Set $\tilde r(g) = \frac{\tilde r(g)}{Lm}$ for $g=1,\cdots,G$.}
	\Classify{$\hat b_x=\argmax_{g\in\{1,\cdots, G\}} \tilde r(g)$.} 
\vspace{0.05 in}
      
\end{algorithm}

\subsection{Computational Complexity}
Given a data point $x\in \mathbb{R}^d$, we require
\[m(\sum_{l=1}^L |T_{l,i}|l+GL)+3G+1\]
flops to predict its class as described in Algorithm~\ref{algo:Class}. Flop counts for each step in Algorithm~\ref{algo:Class} are given in Table~\ref{flops}. We do not include the cost of calculating $q=\text{sign}(Ax)$, as we assume that the algorithm is provided these binary measurements. Additionally, it may be the case that one does not have access to the underlying vector $x$ and only knows the binary measurements $q$.
As the number of levels increases, the term $m\sum_{l=1}^L |T_{l,i}|l$ typically dominates the testing cost. The number of possible sign patterns for a single measurement is $2^l$ and thus we at least have the bound $|T_{l,i}|\le 2^l$. The inequality is strict if not all possible sign patterns are realized by points in the training data. 

\begin{table}
\begin{center}
  \begin{tabular}{ | c | c | c }
    \hline
    \textbf{Flops} & \textbf{Operation}  \\ \hline
    $G$ & Initialize $\tilde r$ \\ \hline
    $m\sum_{l=1}^L |T_{l,i}|l$ & Identify the sign pattern (worst case)  \\ \hline
    $mGL$ & Update $\tilde r(g)$ for each class and level \\ \hline
    $G+1$ & Scale \\ \hline
    $G$ & Predict $\hat b_x=\argmax_{g\in\{1,\cdots, G\}} \tilde r(g)$  \\ \hline
  \end{tabular}
\end{center}
\caption{Testing flop counts.}
\label{flops}
\end{table}

\label{HC_adj}\subsection{Adjustment for Hierarchical Classification}
We now describe our proposed adjustment for handling hierarchical classification, where the labels possess some sort of tree structure. The classification scheme described above and in \cite{NSW17Simple} has the property that more levels (higher $L$) are needed to accurately classify more complicated data. 
Thus, if we know in advance that certain classes may require fewer levels for classification with sufficient accuracy, we may isolate these classes in an initial classification that uses fewer levels and then further classify these groups of classes using only the required number of levels for sufficient accuracy. This strategy leads to computational savings without sacrificing accuracy when some classes are more easily discerned from the others. 

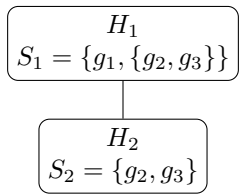
\begin{figure}
\centering
\begin{tikzpicture}[sibling distance=10em,
  every node/.style = {shape=rectangle, rounded corners,
    draw, align=center,
    top color=white, bottom color=white}]]
  \node {$H_1$\\$S_1 = \{g_1,\{g_2,g_3\}\}$}
    child {node {$H_2$\\$S_2 = \{g_2,g_3\}$}} ;
\end{tikzpicture}
\caption{Hierarchical classification tree for a simple three-class example in which differentiating $g_1$ is significantly easier than differentiating $g_2$ and $g_3$.}
\label{fig:2tree}
\end{figure}

For illustration, consider the simple example where we have three classes, $g_1,g_2,g_3$. Suppose that $L_1$ levels are necessary to classify data belonging to $g_1$, but $L_2$ levels are required to differentiate between classes $g_2$ and $g_3$ where $L_2>L_1$. We can perform binary classification between $g_1$ and $\{g_2,g_3\}$ using $L_1$ levels followed by classification between $g_2$ and $g_3$ using $L_2$ levels. These classifications can be organized as a tree with nodes $H_1$ and $H_2$ as shown in Figure \ref{fig:2tree}. The sets $S_1$ and $S_2$ give the class groupings for the model constructed at nodes $H_1$ and $H_2$ respectively.

At test time, points initially classified as $g_1$ require only 
\[m\left(\sum_{l=1}^{L_1} |T_{l,i}|l+G L_1\right)+3G+1\]  flops, where $G=2$. 
In order to further discern between points predicted to belong to $g_2$ or $g_3$, we can use the same measurements (or random hyperplanes) as used in the first classification. Then for the two classifications, points initially classified as belonging to $g_2$ or $g_3$ require 
\[m\left(\sum_{l=1}^{L_2} |T_{l,i}|l+\sum_{c=1}^2 G_c L_c\right)+\sum_{c=1}^2 (3G_c+1)\]
flops to arrive at a prediction. Here, $G_c$ is the number of groups at classification $c$ and $L_c$ is the number of levels used for classification $c$. In this particular example, we would have $G_1=|S_1|=|\{g_1,\{g_2,g_3\}\}|=2$ and $G_2=|S_2|=|\{g_2,g_3\}|=2$. 

The overhead cost to carrying out two classifications instead of one is quite limited overall. For classifications in which some classes require fewer levels to predict, this hierarchical structure can lead to significant computational savings, as shown in the experimental results that follow. The magnitude of the computational savings is highly dependent on the distribution of the testing data, however, as we only reduce computational costs for those points predicted to be in one of the classes that is `easier' to discern, i.e. requires fewer levels. The proposed hierarchical classification algorithm is further described in Algorithms \ref{algo:HCTrain} and \ref{algo:HCTest}.

\begin{algorithm}
    \caption{Proposed adjustment for hierarchical classification (training).}
	\label{algo:HCTrain}
    \SetKwInOut{Input}{input}
    \SetKwInOut{Init}{initialize}
    \SetKwInOut{Id}{identify}
    \SetKwInOut{Update}{update}
    \SetKwInOut{Scale}{scale}
    \SetKwInOut{Classify}{classify}
    \SetKwInOut{Train}{train}
    \SetKwInOut{Define}{define}
    \SetKwInOut{Set}{set}

    \Input{binary training data $Q$, training labels $b$, set of class groupings $S_c$ for each node $H_c$ in the tree of classifications $H$, number of levels $L = (L_1,\cdots L_C)$ to be used in each classification.}
    \For{$H_c\in H$, }
      {
	\Id{Identify rows of $Q$ such that the corresponding component of $b$ is in one of the sets contained in $S_c$. Form a matrix $Q_c$ containing these rows.}
	\Define{Define $\tilde b$ to be the labels indicating to which set of $S_c$ a given row of $Q_c$ corresponds.}
       \Train{Train a classifier as in Algorithm \ref{algo:Train} 
with training data $Q_c$, labels $\tilde b$, number of groups $|S_c|$ and number of levels $L_c$ as input.}
	}
\vspace{0.05 in}
  
\end{algorithm}

\begin{algorithm}
    \caption{Proposed adjustment for hierarchical classification (testing).}
	\label{algo:HCTest}
    \SetKwInOut{Input}{input}
    \SetKwInOut{Init}{initialize}
    \SetKwInOut{Id}{identify}
    \SetKwInOut{Update}{update}
    \SetKwInOut{Scale}{scale}
    \SetKwInOut{Predict}{predict}
    \SetKwInOut{Classify}{classify}
    \SetKwInOut{Begin}{begin}
    \SetKwInOut{Define}{define}
    \SetKwInOut{Set}{set}

    \Input{binary testing data $q$, set of class groupings $S_c$, learned parameters $r(l,i,t,g),$ $T_{l,i}$ and $\Lambda_{l,i}$ for the classification associated to each node $H_c$ in the tree of classifications $H$, number of levels $L = (L_1,\cdots L_C)$ to be used in each classification.}
	\Begin{Begin at $H_1$, the root classification.}
    \While{$H_c$ is not null, }
      {
	\Classify{Classify testing data $q$, as in Algorithm \ref{algo:Class}, with learned parameters $r(l,i,t,g)$, $T_{l,i}$, $\Lambda_{l,i}$ from $H_c$ into one of the sets contained in $S_c$. }  
	\uIf{If $q$ is predicted to belong to a single class}{
	\Set{Set $H_c$ to be null.}
	}
	\Else{ \Set{Set $H_c$ to be the node corresponding to the predicted set of classes for $q$ within $S_c$.}}
	}
\vspace{0.05 in}
  
\end{algorithm}

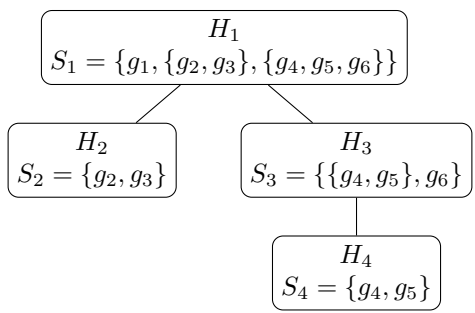
\begin{figure}
\centering
\begin{tikzpicture}[sibling distance=10em,
  every node/.style = {shape=rectangle, rounded corners,
    draw, align=center,
    top color=white, bottom color=white}]]
  \node {$H_1$\\$S_1 = \{g_1,\{g_2,g_3\},\{g_4,g_5,g_6\}\}$}
    child {node {$H_2$\\$S_2 = \{g_2,g_3\}$}}
    child { node {$H_3$\\$S_3 = \{\{g_4,g_5\},g_6\}$}
      child { node {$H_4$\\$S_4 = \{g_4,g_5\}$}} };
\end{tikzpicture}
\caption{Example hierarchical classification tree. A classifier would be trained at each node, $H_c$, to classify data among the sets given in $S_c$.}
\label{fig:tree}
\end{figure}

This hierarchical classification strategy naturally generalizes to incorporate more complicated and deeper hierarchical structures in which the classifications can be structured as a tree. See Figure \ref{fig:tree} for an example.  In order to maximize computational gains, however, we would like the tree to be 	`imbalanced' in terms of the maximum number of levels required for sufficient  classification accuracy along different paths of the tree. Such an imbalance arises naturally in many applications. For example, consider brain imaging and the problem of detecting brain abnormalities including tumors and dementia; tumor detection is a fairly easy learning problem whereas classifying differing types of dementia remains very challenging \cite{duncan2016classification,higdon2004comparison}.  

\subsection{Determining class hierarchies}
When the number of classes is large, reorganizing a flat multiclass classification problem into hierarchical (binary) classifications can be used as a general strategy to reduce the computation required for testing \cite{griffin2008learning}. Our proposed strategy need not only be applied in settings where the data follows or is presented within the context of a clear hierarchical structure.  A variety of previous works have studied ways in which to detect structure among classes and use this information to construct an informed hierarchy of the classes \cite{griffin2008learning,godbole2002scaling,silva2017improving,li2007hierarchical,zupan1999learning}.  These strategies generally aim to group classes that are deemed `similar' by some measure, in order to reduce the number of missclassifications that occur high in the tree. For example, some work suggests constructing a hierarchy based on the confusion matrix of the flat multiclass classification problem \cite{griffin2008learning,godbole2002scaling,silva2017improving}.
Preferentially constructing class hierarchies that are imbalanced in terms of ease of classification along different paths will also largely affect the computational savings achieved by our proposed hierarchical classification method. We save details of how one might achieve this for future work.

\section{Experimental results}
In the following experiments, we test the computational gains achieved by the proposed hierarchical classification strategy as described in Algorithms \ref{algo:HCTrain} and \ref{algo:HCTest} compared with direct classification into each individual group via `flat multiclass classification' as described in Algorithms \ref{algo:Train} and \ref{algo:Class}. The `flat multiclass classification' is a direct application of the method proposed in \cite{NSW17Simple}.

\subsection{Two-dimensional synthetic data}
\begin{figure}[h]
\centering
\includegraphics[width=0.45\textwidth]{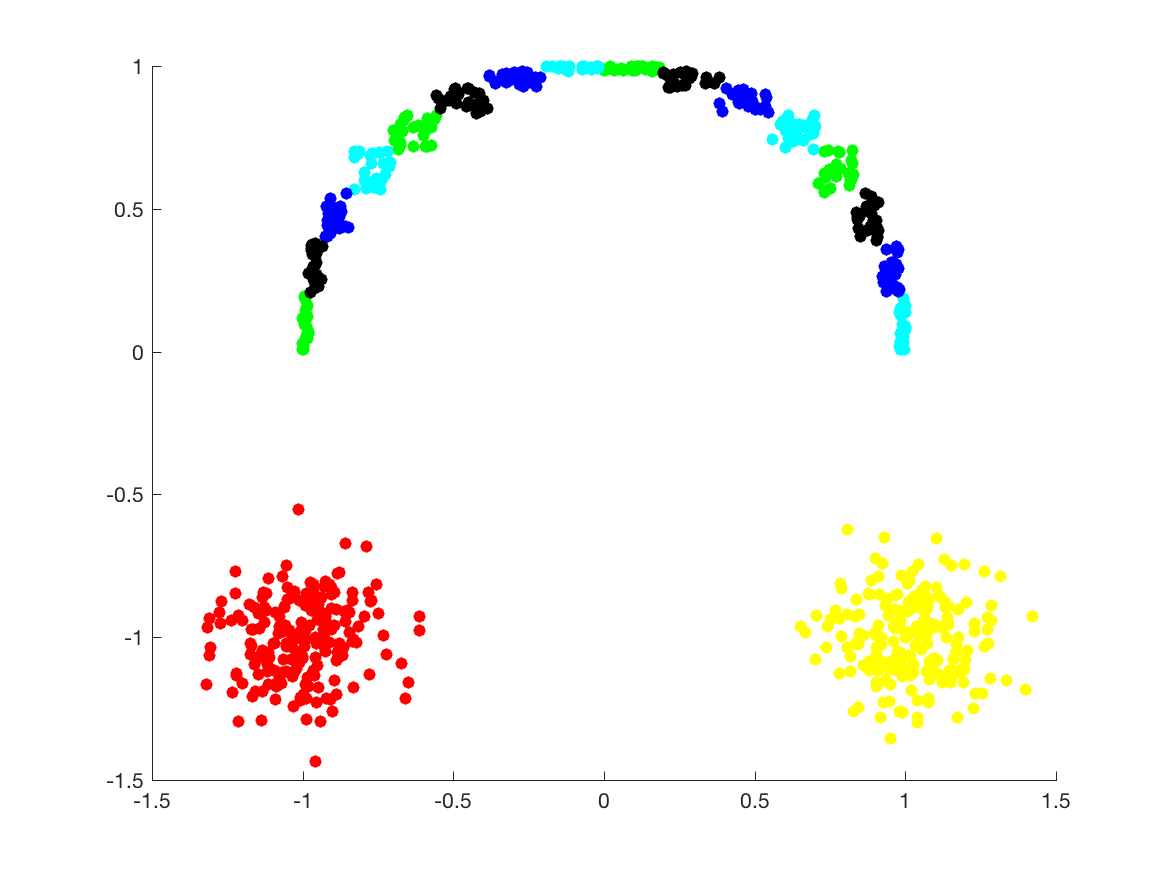}
\includegraphics[width=0.45\textwidth]{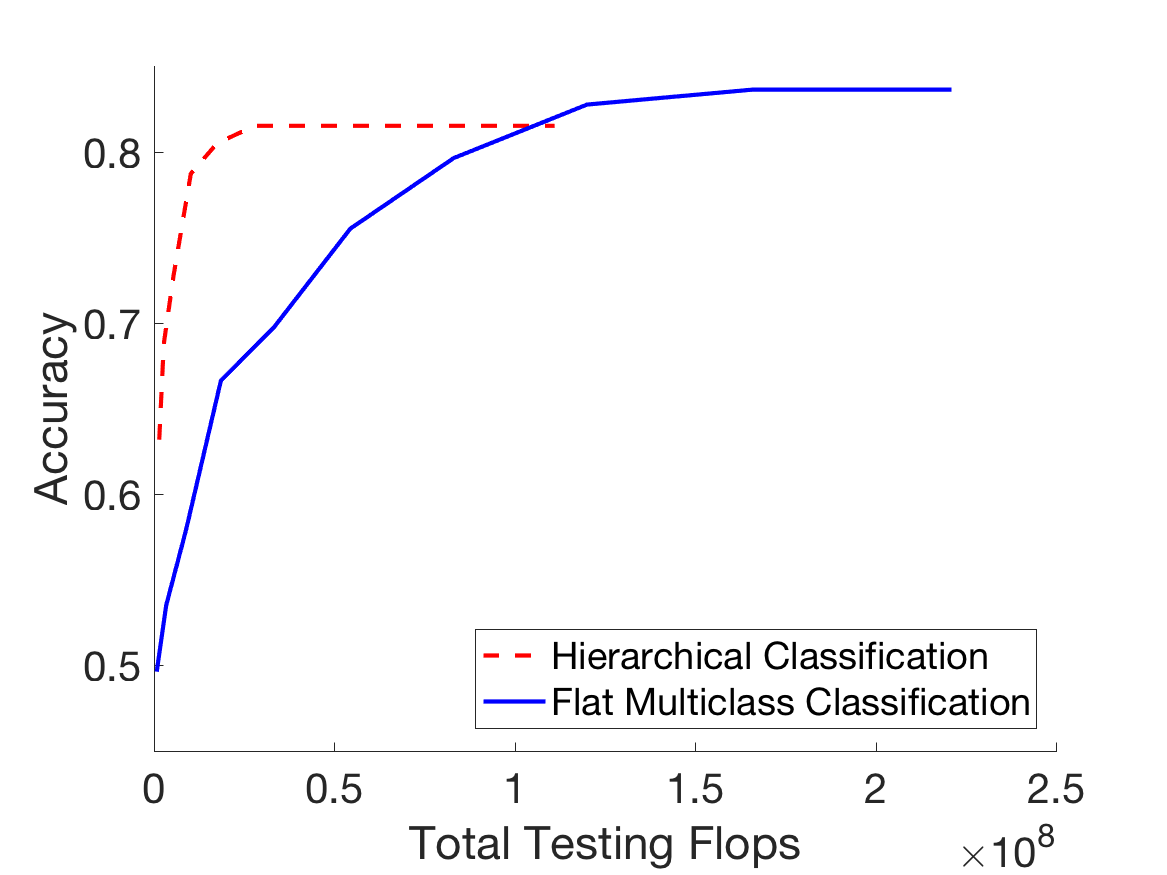}
\includegraphics[width=0.45\textwidth]{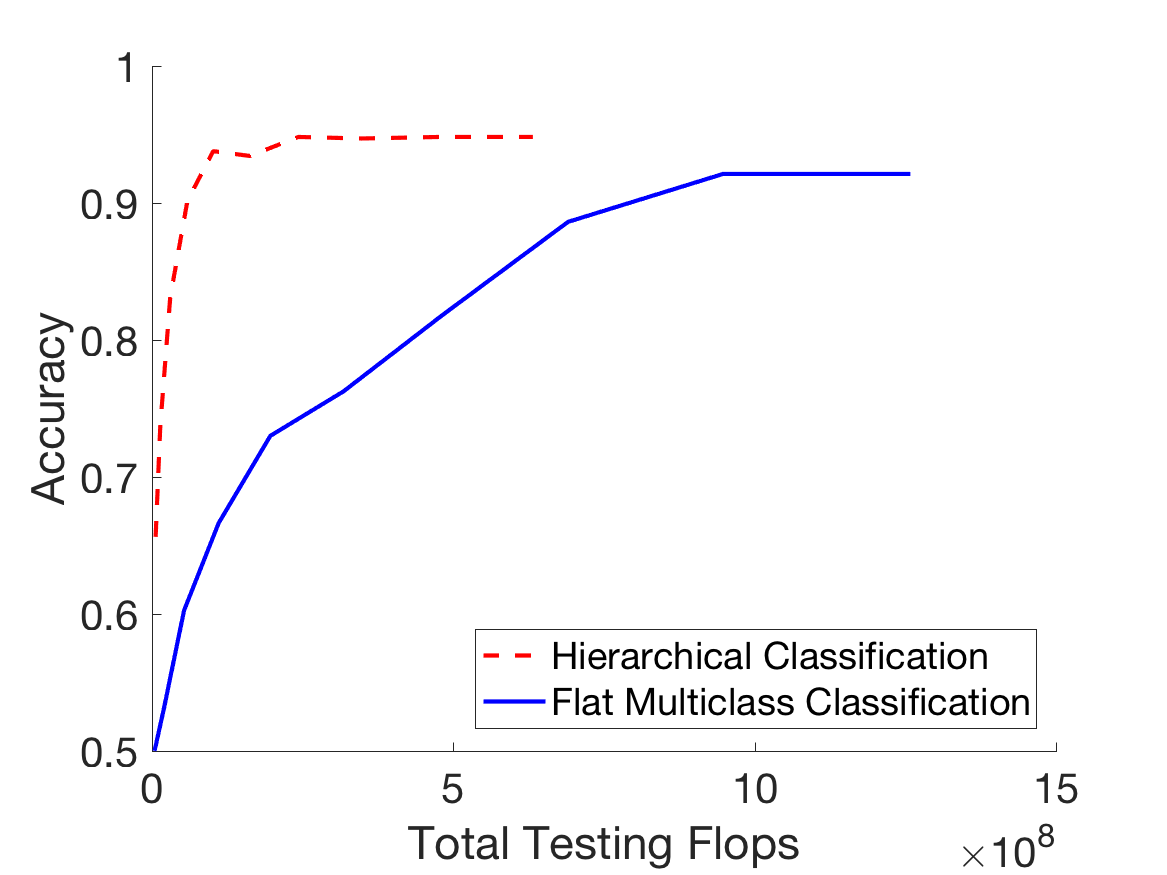}
\includegraphics[width=0.45\textwidth]{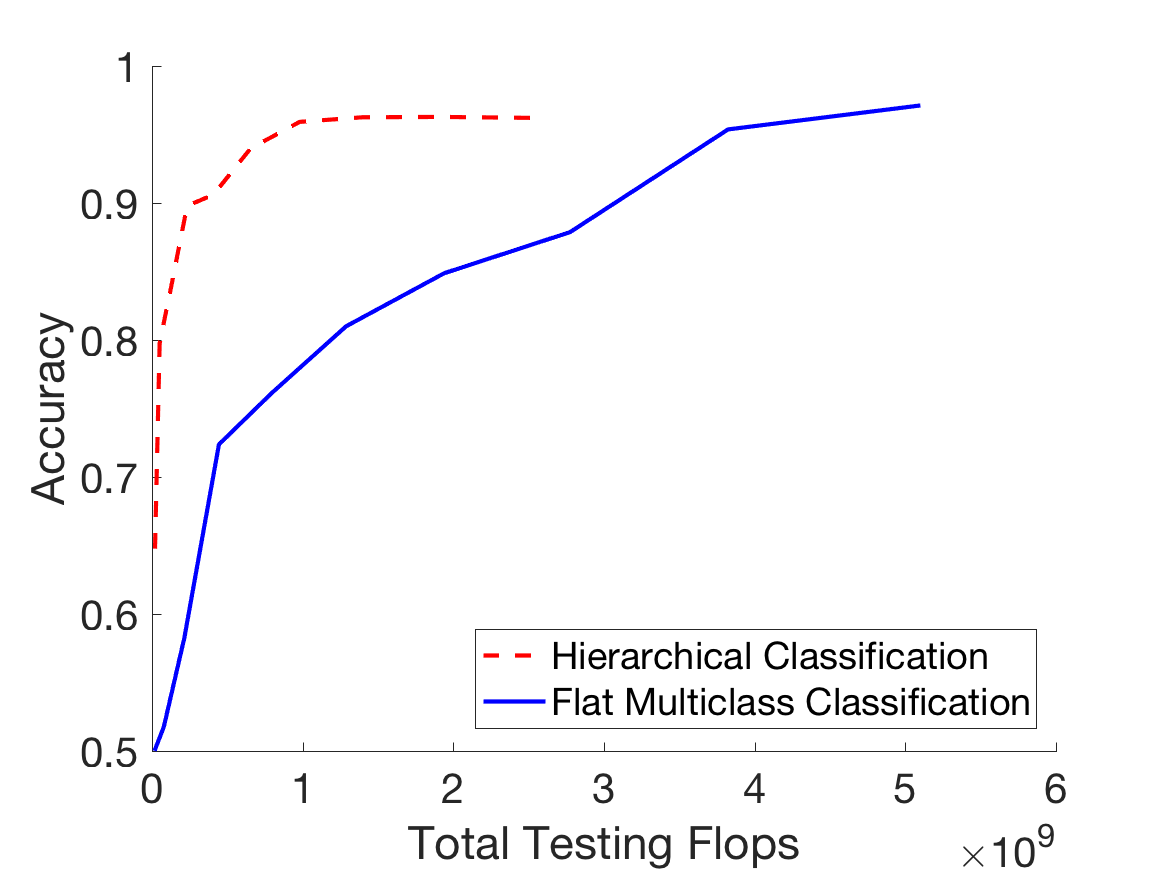}
\caption{For the data distributed as given in the upper-left plot, where each color represents a different class, we classify testing data either by flat multiclass classification or our proposed hierarchical classification strategy where the first classification discerns between red or yellow versus green, black, blue or cyan. Accuracy and testing flops required are given in the subsequent plots using $m=20,50$ and $100$ respectively. Results are averaged over 10 trials.}
\label{fig:2D}
\end{figure}

We first test the computational gains achieved by the proposed hierarchical classification strategy on the two-dimensional data shown in Figure~\ref{fig:2D}. Each color represents a different class and there are six classes in total. The red and yellow clusters each contain 200 training and testing points, while the remaining four classes, green, black, blue and cyan, contain 100 training and testing points each. The distribution of testing points among the classes will have a significant effect on the computation needed for testing in the hierarchical case. We expect classifying points from the red and yellow classes to be easier and to require fewer levels than correctly classifying points as green, black, blue or cyan. 

To take advantage of this structure in the data, we first predict whether a testing point is red or yellow versus green, black, blue or cyan using only one level. If the test point was predicted to be red or yellow, we then discern between these two classes again using only a single level. If the test point was predicted to be green, black, blue or cyan, we then predict among these classes by using varying numbers of levels.  Accuracies and testing flops for the hierarchical classification strategy versus flat multiclass classification are shown in Figure~\ref{fig:2D} for varying numbers of measurements $m$. We see a significant reduction in computational cost using the hierarchical strategy without sacrificing accuracy. 

\begin{figure}[h]
\centering
\begin{tikzpicture}[sibling distance=12em,
  every node/.style = {shape=rectangle, rounded corners,
    draw, align=center,
    top color=white, bottom color=white}]]
  \node {$H_1$\\$S_1 = \{\{\text{red, yellow}\},\{\text{green, black, blue, cyan}\}\}$}
    child {node {$H_2$\\$S_2 = \{\text{red, yellow}\}$}}
    child { node {$H_3$\\$S_3 = \{\text{green, black, blue, cyan}\}$}
      };
\end{tikzpicture}
\caption{Hierarchical classification tree used to classify two-dimensional synthetic data as shown in Figure \ref{fig:2D}.} 
\label{fig:2Dtree}
\end{figure}
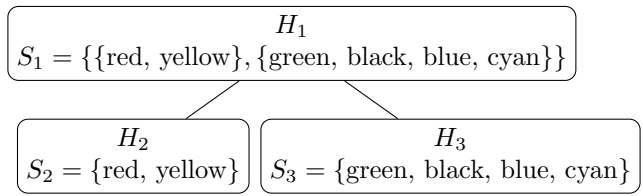

\subsection{Three-dimensional synthetic data}
\begin{figure}[h]
\centering
\includegraphics[width=0.45\textwidth]{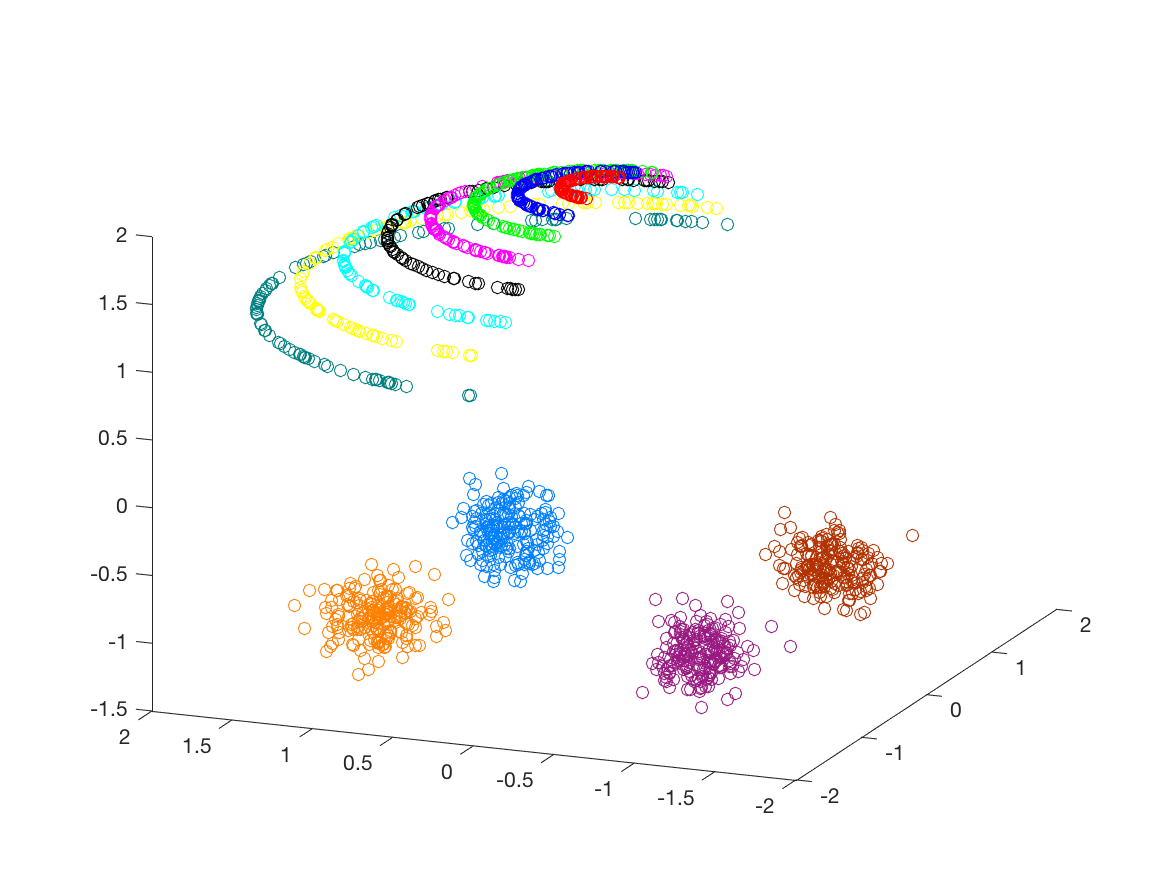}
\includegraphics[width=0.45\textwidth]{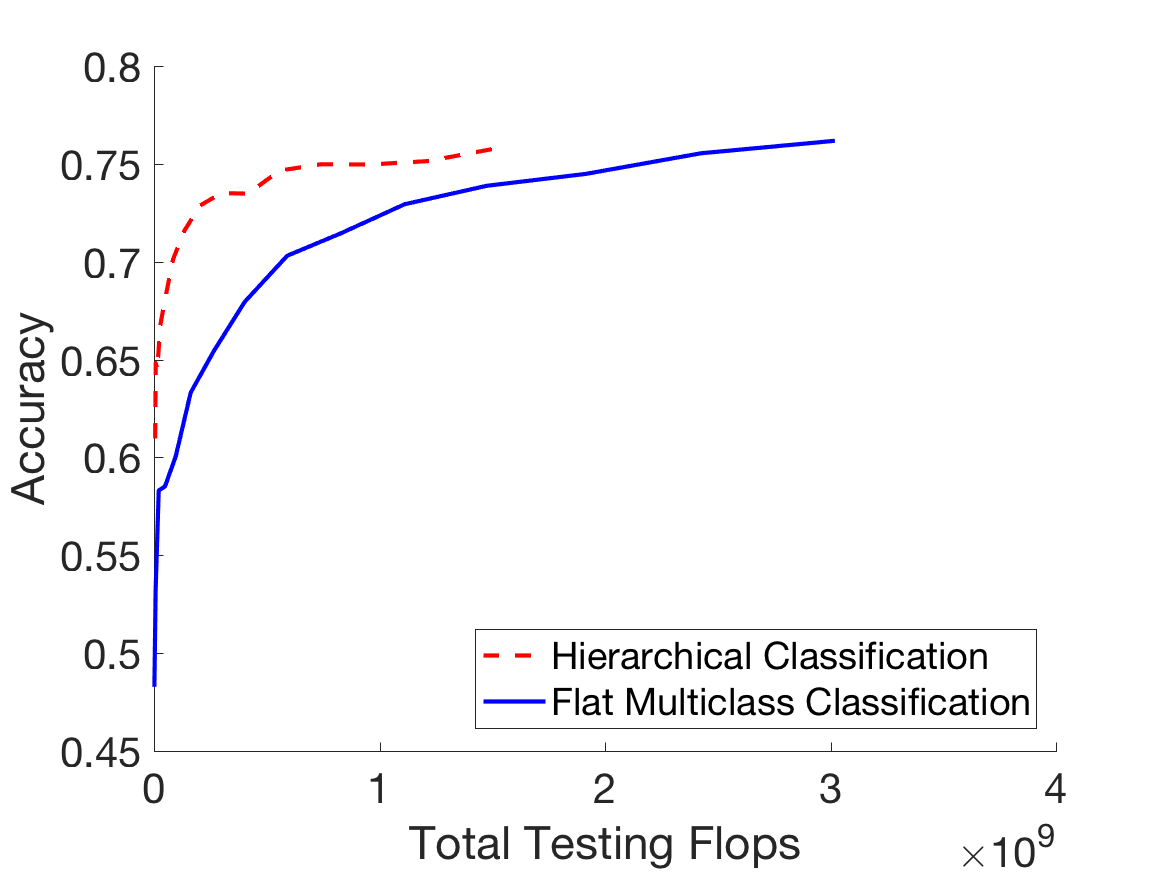}
\includegraphics[width=0.45\textwidth]{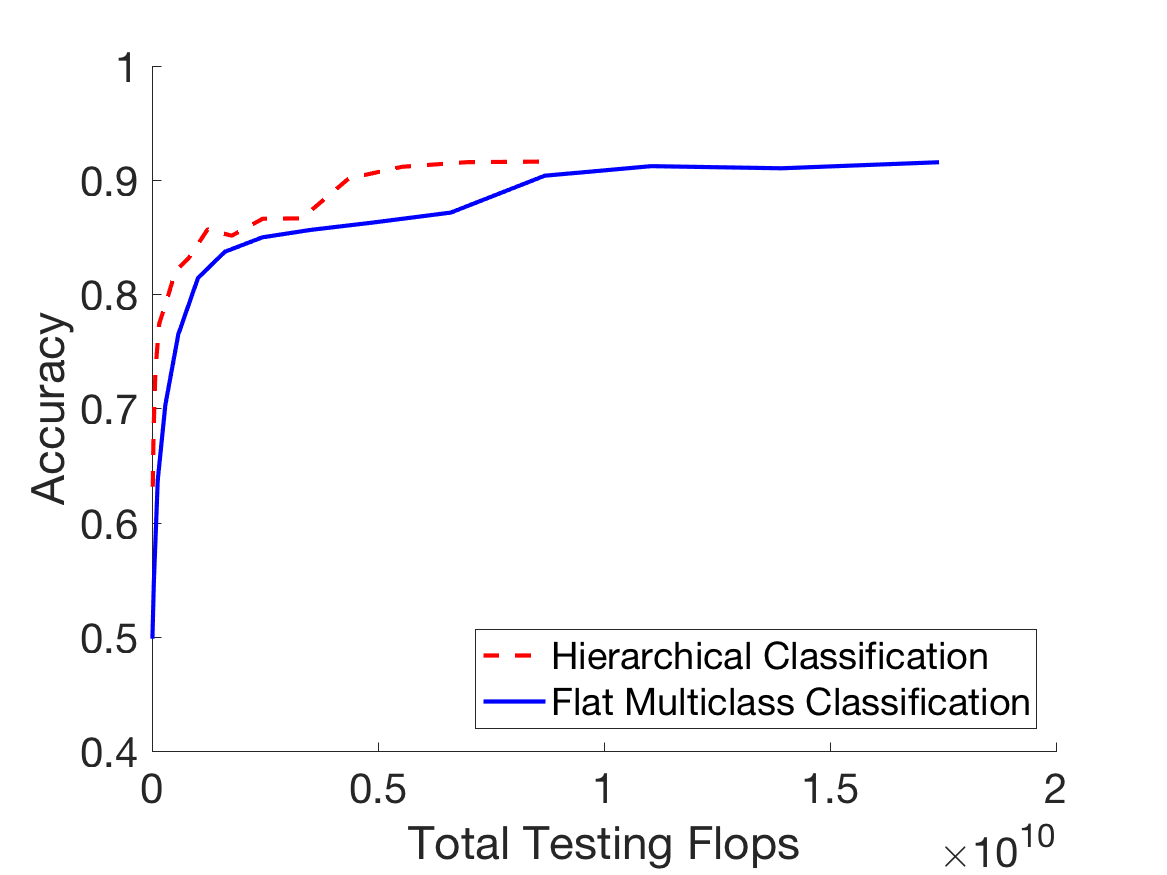}
\includegraphics[width=0.45\textwidth]{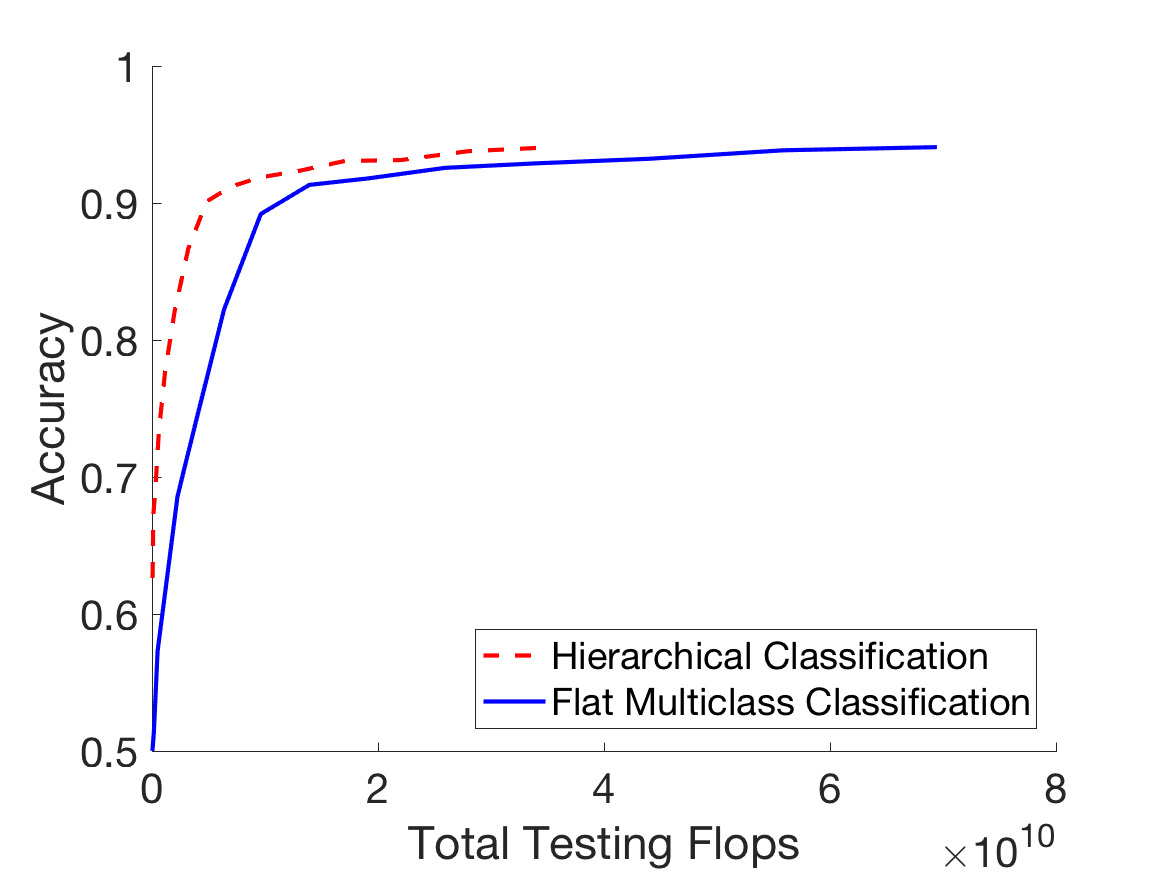}
\caption{For the data distributed as given in the upper-left plot, where each color represents a different class, we classify testing data either by flat multiclass classification or our proposed hierarchical classification strategy where the first classification discerns between red or yellow versus green, black, blue or cyan. Accuracy and testing flops required are given in the subsequent plots using $m=20,50$ and $100$ respectively. Results are averaged over 10 trials.}
\label{fig:3D}
\end{figure}

We test the hierarchical classification strategy and flat multiclass classification on three-dimensional synthetic data as given in Figure~\ref{fig:3D}. Each color represents a different class. Again, we expect the four Gaussian clusters to require fewer levels for sufficiently accurate classification than the `arcs'. The training data are distributed so that there is an equal number of training and testing points in the Gaussian clusters and arcs. Specifically we have 100 training and testing points in each arc and 200 training and testing points in each Gaussian cluster. 

Using a strategy similar to that used in the two-dimensional experiment, we first build a classifier to predict whether a point belongs to one of the arcs or one of the Gaussian clusters using only a single level. If a data point is predicted to be in one of the Gaussian clusters, we then use a single level again to predict to which of the clusters it belongs. If a data point is predicted to be in one of the arcs, we use more levels to perform the subsequent classification to discern between the arcs. We test the accuracy and computation required for using a variety of levels in this second classification. As in the two-dimensional experiment, we again see a reduction in the computational cost of testing without sacrificing accuracy.

\subsection{MNIST}
Although not inherently hierarchical in nature, we demonstrate that our hierarchical strategy can lead to computational savings on the MNIST dataset of handwritten digits \cite{MNIST}. Consider the digits 1-5. Intuitively and in practice, the digit 1 tends to be easier to classify correctly than the other digits. For example, if we apply the multiclass classification from \cite{NSW17Simple} to classify the digits 1-5 using 1000 training points for each class, 10 levels and testing on 200 training points from each class, we find that 98.5\% of the 1s are classified correctly, whereas the overall accuracy of classifying the digits 1-5 was 89.2\% (the accuracy for classifying digits 2-5 was 86.88\%). Thus, it is reasonable to expect that fewer levels are required for sufficiently accurate classification of the 1s than are required to classify the remaining digits.

We induce hierarchical structure by first classifying into 1s versus not 1s, followed by classification into the digits 2,3,4 and 5 for those test points that were predicted to not be 1s in the first classification. When training the first classifier, we downsample the training data for the digits 2-5 so that we have an equal number of training data points for 1s and not 1s. We found that this adjustment improved the accuracy of the first classification. Five levels are used for the first classification into 1s versus not 1s and a varying number of levels (five to 10) are used for the subsequent classification. We again see a reduction in the total testing flops required to achieve a given accuracy. Here, we use an equal number of test points for each digit and thus get computational savings for approximately 1/5 of the test points, specifically for all of the test points that are predicted to be 1s. If we had a much higher proportion of 1s as compared to the other digits, then we would expect the computational savings to be even more significant. Additionally, since this tree is fairly shallow, as expected the improvements are mild, and we would expect more significant improvement for real data that has a larger and more imbalanced tree structure, as in the other experiments.

\begin{figure}[h]
\centering
\includegraphics[width=0.45\textwidth]{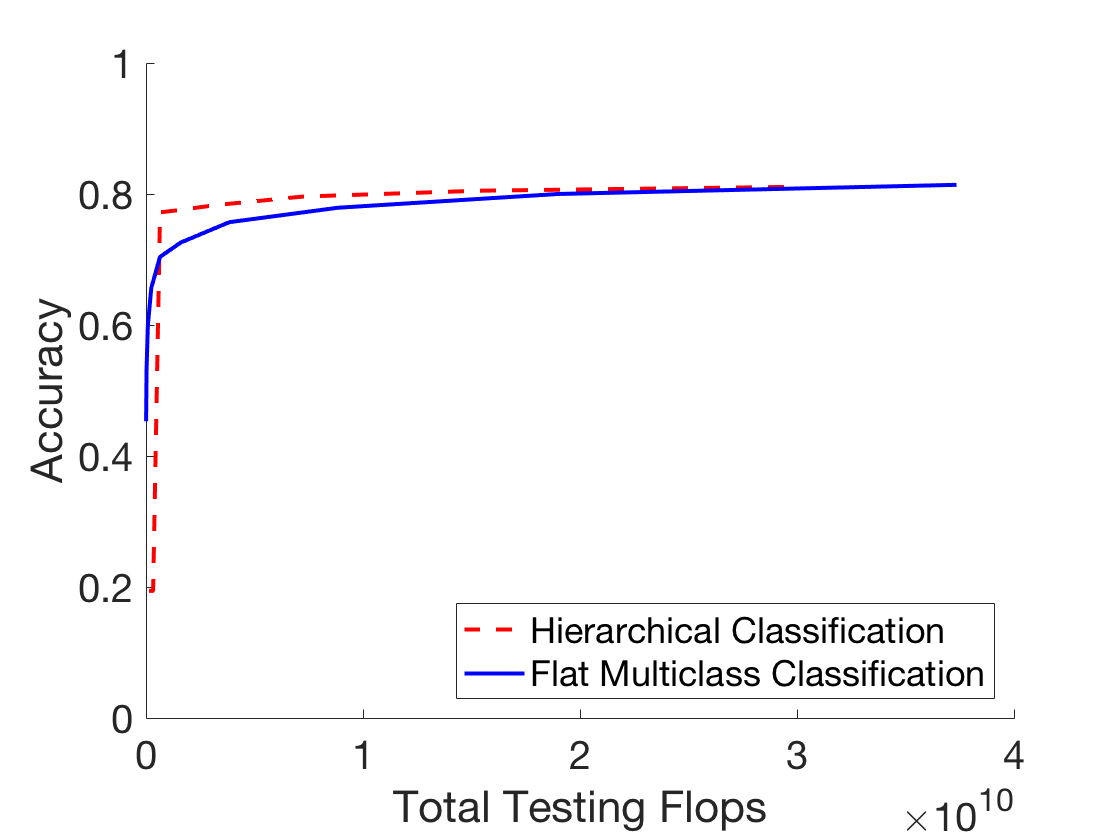}
\includegraphics[width=0.45\textwidth]{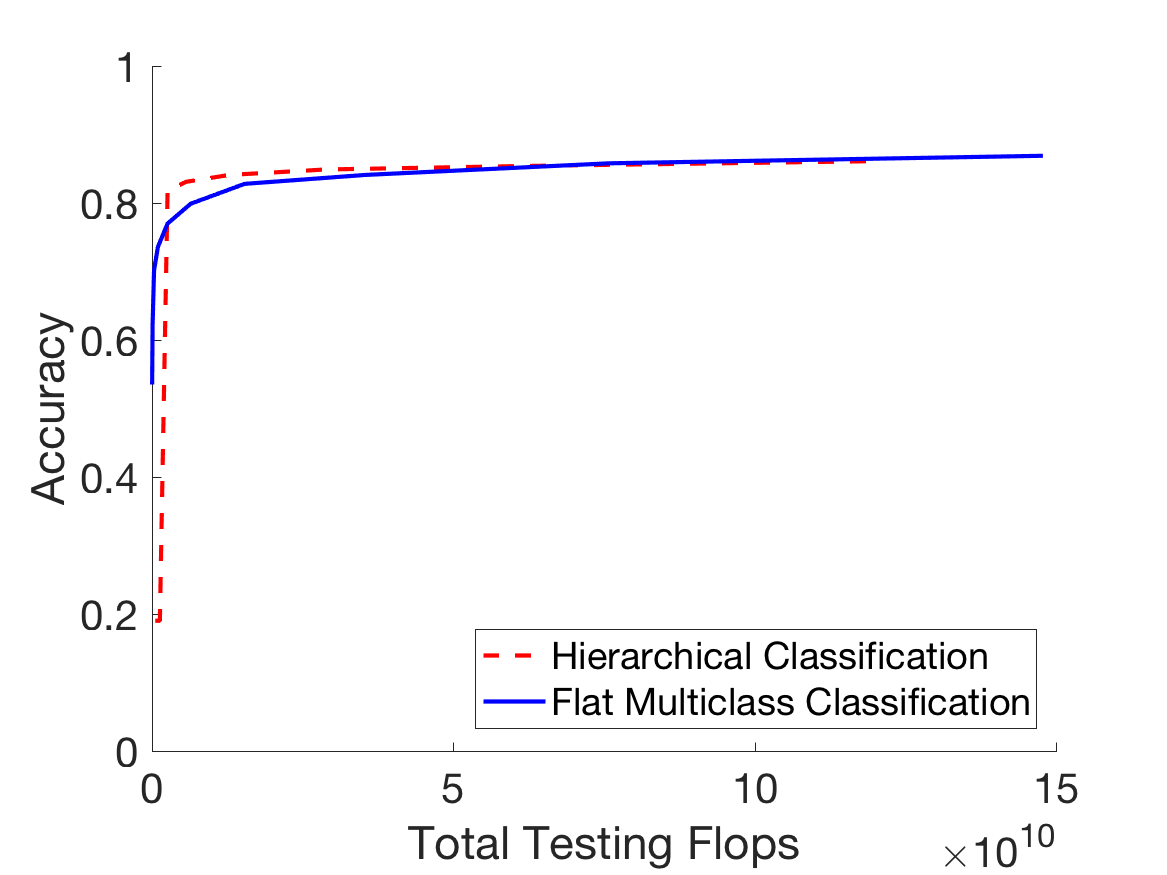}
\includegraphics[width=0.45\textwidth]{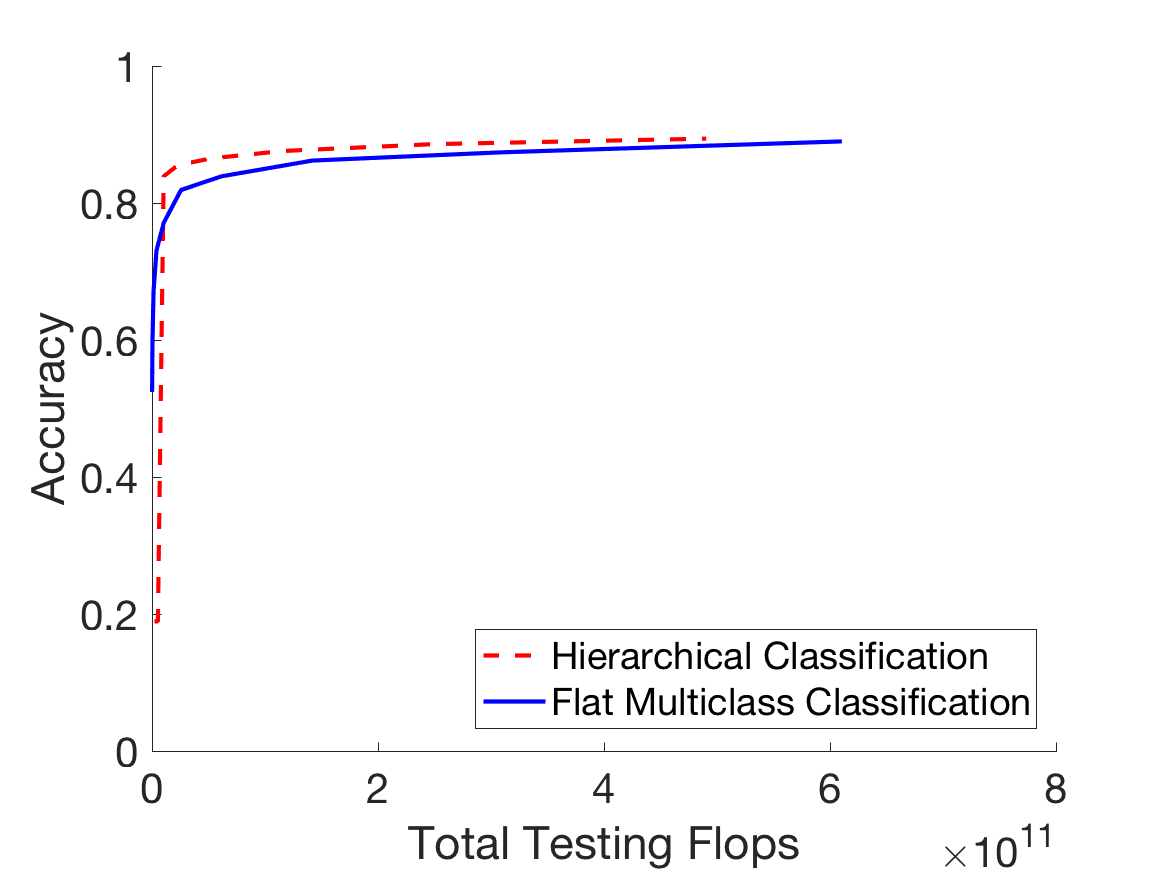}
\includegraphics[width=0.45\textwidth]{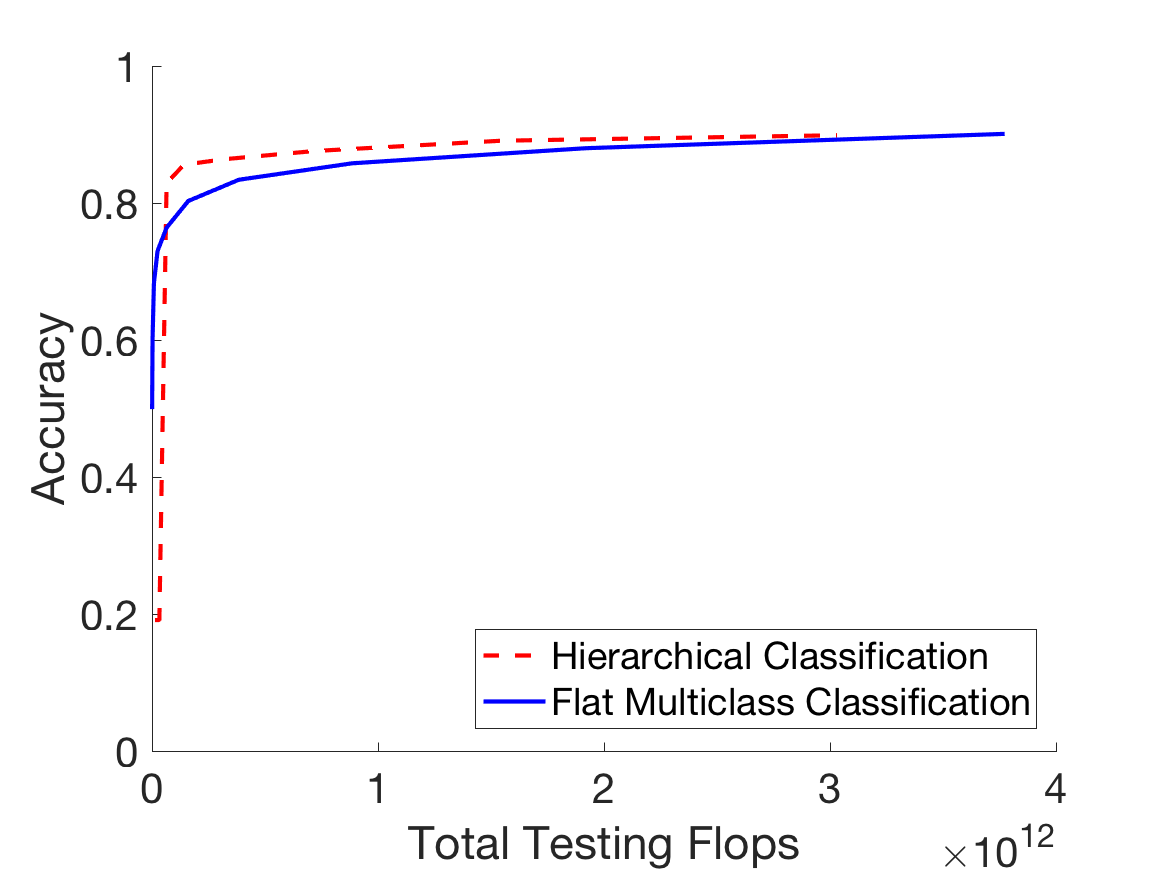}
\caption{Accuracy and testing flops required for flat multiclass classification versus our proposed hierarchical classification strategy in classifying digits 1-5 in the MNIST dataset are given using $m=50, 100, 200$ and $500$ respectively. Results are averaged over 10 trials.}
\label{fig:MNIST}
\end{figure}

\section*{Conclusion}
We have demonstrated that the classification algorithm proposed in \cite{NSW17Simple} can be readily adapted to to classify data in a hierarchical way that improves computational efficiency. We achieve this by using fewer levels to classify data points predicted to be from classes that are more readily identifiable. 
We could potentially further reduce computational costs for easier to classify data by reducing the number of measurements $m$ in those cases as well. Theoretical guarantees as well as modifications that alleviate error propagation down the tree are important directions for future work.

\section*{Acknowledgments}
Molitor and Needell were partially supported by NSF CAREER grant $\#1348721$ and NSF BIGDATA $\#1740325$.

\bibliographystyle{plain}
\bibliography{dnbib}

\begin{thebibliography}{10}

\bibitem{aziz1996overview}
Pervez~M Aziz, Henrik~V Sorensen, and J~Vn~der Spiegel.
\newblock An overview of sigma-delta converters.
\newblock {\em IEEE signal processing magazine}, 13(1):61--84, 1996.

\bibitem{RefWorks:526}
L.~Bottou and O.~Bousquet.
\newblock The tradeoffs of large-scale learning.
\newblock {\em Optimization for Machine Learning}, page 351, 2011.

\bibitem{cheong2004support}
Sungmoon Cheong, Sang~Hoon Oh, and Soo-Young Lee.
\newblock Support vector machines with binary tree architecture for multi-class
  classification.
\newblock {\em Neural Information Processing-Letters and Reviews}, 2(3):47--51,
  2004.

\bibitem{duncan2016classification}
Dominique Duncan and Thomas Strohmer.
\newblock Classification of alzheimer s disease using unsupervised diffusion
  component analysis.
\newblock {\em Mathematical Biosciences and Engineering}, 13:1119--1130, 2016.

\bibitem{fang2014sparse}
Jun Fang, Yanning Shen, Hongbin Li, and Zhi Ren.
\newblock Sparse signal recovery from one-bit quantized data: An iterative
  reweighted algorithm.
\newblock {\em Signal Processing}, 102:201--206, 2014.

\bibitem{godbole2002scaling}
Shantanu Godbole, Sunita Sarawagi, and Soumen Chakrabarti.
\newblock Scaling multi-class support vector machines using inter-class
  confusion.
\newblock In {\em Proceedings of the eighth ACM SIGKDD international conference
  on Knowledge discovery and data mining}, pages 513--518. ACM, 2002.

\bibitem{gordon1987review}
Allan~D Gordon.
\newblock A review of hierarchical classification.
\newblock {\em Journal of the Royal Statistical Society. Series A (General)},
  pages 119--137, 1987.

\bibitem{griffin2008learning}
Gregory Griffin and Pietro Perona.
\newblock Learning and using taxonomies for fast visual categorization.
\newblock In {\em Computer Vision and Pattern Recognition, 2008. CVPR 2008.
  IEEE Conference on}, pages 1--8. IEEE, 2008.

\bibitem{RefWorks:439}
A.~Gupta, R.~Nowak, and B.~Recht.
\newblock Sample complexity for 1-bit compressed sensing and sparse
  classification.
\newblock In {\em International Symposium on Information Theory (ISIT)}. IEEE,
  2010.

\bibitem{higdon2004comparison}
Roger Higdon, Norman~L Foster, Robert~A Koeppe, Charles~S DeCarli, William~J
  Jagust, Christopher~M Clark, Nancy~R Barbas, Steven~E Arnold, R~Scott Turner,
  Judith~L Heidebrink, et~al.
\newblock A comparison of classification methods for differentiating
  fronto-temporal dementia from alzheimer's disease using fdg-pet imaging.
\newblock {\em Statistics in medicine}, 23(2):315--326, 2004.

\bibitem{biht}
L.~Jacques, J.~N. Laska, P.~T. Boufounos, and R.~G. Baraniuk.
\newblock Robust 1-bit compressive sensing via binary stable embeddings of
  sparse vectors.
\newblock {\em IEEE T. Inform. Theory}, 59(4):2082--2102, 2011.

\bibitem{MNIST}
Yann LeCun, Corinna Cortes, and CJ~Burges.
\newblock Mnist handwritten digit database.
\newblock {\em AT\&T Labs [Online]. Available: http://yann. lecun.
  com/exdb/mnist}, 2, 2010.

\bibitem{li2007hierarchical}
Tao Li, Shenghuo Zhu, and Mitsunori Ogihara.
\newblock Hierarchical document classification using automatically generated
  hierarchy.
\newblock {\em Journal of Intelligent Information Systems}, 29(2):211--230,
  2007.

\bibitem{NSW17Simple}
D.~Needell, R.~Saab, and T.~Woolf.
\newblock Simple classification using binary data.
\newblock 2017.
\newblock Submitted.

\bibitem{silla2011survey}
Carlos~N Silla and Alex~A Freitas.
\newblock A survey of hierarchical classification across different application
  domains.
\newblock {\em Data Mining and Knowledge Discovery}, 22(1-2):31--72, 2011.

\bibitem{silva2017improving}
Daniel Silva-Palacios, C{\`e}sar Ferri, and Mar{\'\i}a~Jos{\'e}
  Ram{\'\i}rez-Quintana.
\newblock Improving performance of multiclass classification by inducing class
  hierarchies.
\newblock {\em Procedia Computer Science}, 108:1692--1701, 2017.

\bibitem{weston1998multi}
Jason Weston and Chris Watkins.
\newblock Multi-class support vector machines.
\newblock Technical report, Citeseer, 1998.

\bibitem{zupan1999learning}
Bla{\v{z}} Zupan, Marko Bohanec, Janez Dem{\v{s}}ar, and Ivan Bratko.
\newblock Learning by discovering concept hierarchies.
\newblock {\em Artificial Intelligence}, 109(1-2):211--242, 1999.

\end{thebibliography}

\end{document}